\documentclass[10pt,twocolumn,letterpaper]{article}

\usepackage{cvpr}
\usepackage{times}
\usepackage{epsfig}
\usepackage{graphicx}

\usepackage{amssymb}
\usepackage{amsfonts}
\usepackage{mathtools}
\usepackage{cite}
\usepackage{amsthm}
\usepackage{algorithm}
\usepackage{algpseudocode}
\usepackage{bm}
\usepackage{cases}
\usepackage{amsmath}

\makeatletter
\newcommand{\printfnsymbol}[1]{%
  \textsuperscript{\@fnsymbol{#1}}%
}
\makeatother

\newtheorem{theorem}{Theorem}
\newtheorem{lemma}{Lemma}

\newcommand{\R}{\mathbb{R}}
\newcommand{\1}{\mathbf{1}}
\newcommand{\0}{\mathbf{0}}
\newcommand{\p}{\mathbf{p}}
\newcommand{\g}{\mathbf{g}}
\newcommand{\z}{\mathbf{z}}
\newcommand{\uu}{\mathbf{u}}
\newcommand{\uup}{\mathbf{u}^\prime}
\newcommand{\muu}{\bm{\mu}}

\newcommand{\soft}{\mathrm{softmax}}
\newcommand{\sparse}{\mathrm{sparsemax}}
\newcommand{\spest}{\mathrm{SparsestMax}}
\newcommand*{\tran}{{^{\mkern-1.5mu\mathsf{T}}}}

\usepackage[pagebackref=true,breaklinks=true,letterpaper=true,colorlinks,bookmarks=false]{hyperref}

\cvprfinalcopy 

\def\cvprPaperID{778} 

\ifcvprfinal\fi

\begin{document}

\title{SSN: Learning Sparse Switchable Normalization via SparsestMax}

\author{
Wenqi Shao\textsuperscript{1}\thanks{equal contribution. This work will be presented at CVPR 2019.}\\
\tt\small{weqish@link.cuhk.edu.hk}
\and
Tianjian Meng\textsuperscript{2}\printfnsymbol{1}\\
\tt\small{tianjian.meng@pitt.edu}
\and
Jingyu Li\textsuperscript{2}\\
\tt\small{lijingyu@sensetime.com}
\and
Ruimao Zhang\textsuperscript{1}\\
\tt\small{ruimao.zhang@cuhk.edu.hk}
\and
Yudian Li\textsuperscript{2}\\
\tt\small{liyudian@sensetime.com}
\and
Xiaogang Wang\textsuperscript{1}\\
\tt\small{xgwang@ee.cuhk.edu.hk}
\and
Ping Luo\textsuperscript{1}\\
\tt\small{pluo.lhi@gmail.com}\\
{\textsuperscript{1} The Chinese University of Hong Kong ~~ \textsuperscript{2} SenseTime Research}}

\maketitle
\begin{abstract}
Normalization methods improve both optimization and generalization of ConvNets. To further boost performance, the recently-proposed switchable normalization (SN) provides a new perspective for deep learning: it learns to select different normalizers for different convolution layers of a ConvNet. However, SN uses softmax function to learn importance ratios to combine normalizers, leading to redundant computations compared to a single normalizer.

This work addresses this issue by presenting Sparse Switchable Normalization (SSN) where the importance ratios are constrained to be sparse. Unlike $\ell_1$ and $\ell_0$ constraints that impose difficulties in optimization, we turn this constrained optimization problem into feed-forward computation by proposing SparsestMax, which is a sparse version of softmax. SSN has several appealing properties. (1) It inherits all benefits from SN such as applicability in various tasks and robustness to a wide range of batch sizes. (2) It is guaranteed to select only one normalizer for each normalization layer, avoiding redundant computations. (3) SSN can be transferred to various tasks in an end-to-end manner. Extensive experiments show that SSN outperforms its counterparts on various challenging benchmarks such as ImageNet, Cityscapes, ADE20K, and Kinetics.

\end{abstract}
\section{Introduction}

Normalization techniques~\cite{ba2016layer,ioffe2015batch,ulyanov2017instance,wu2018group} such as batch normalization (BN) \cite{ioffe2015batch} are indispensable components in deep neural networks (DNNs) \cite{he2016deep,huang2017densely}. They improve both learning and generalization capacity of DNNs.
%
Different normalizers have different properties.
For example, BN~\cite{ioffe2015batch} acts as a regularizer and improves generalization of a deep network \cite{luo2018understanding}.
Layer normalization (LN)~\cite{ba2016layer} accelerates the training of recurrent neural networks (RNNs) by stabilizing the hidden states in them.
Instance normalization (IN)~\cite{ulyanov2017instance} is able to filter out complex appearance variances~\cite{pan2018two}.
Group normalization (GN)~\cite{wu2018group} achieves stable accuracy in a wide range of batch sizes.

To further boost performance of DNNs, the recently-proposed Switchable Normalization (SN)
\cite{luo2018differentiable} offers a new viewpoint in deep learning: it learns importance ratios to compute the weighted average statistics of IN, BN and LN, so as to learn different combined normalizers for different convolution layers of a DNN.
%
SN is applicable in various computer vision problems and robust to a wide range of batch sizes.
Although SN has great successes,
it suffers from slowing the testing speed because each normalization layer is a combination of multiple normalizers.
%

To address the above issue, this work proposes \textit{Sparse Switchable Normalization} (SSN) that learns to select a single normalizer from a set of normalization methods for each convolution layer.
%
%
Instead of using $\ell_1$ and $\ell_0$ regularization to learn such sparse selection, which increases the difficulty of training deep networks, SSN turns this constrained optimization problem into feed-forward computations, making auto-differentiation applicable in most popular deep learning frameworks to train deep models with sparse constraints in an end-to-end manner.

In general, this work has three main \textbf{contributions}.

(1) We present Sparse Switchable Normalization (SSN) that learns to select a single normalizer for each normalization layer of a deep network to improve generalization ability and speed up inference compared to SN.
%
SSN inherits all advantages from SN, for example, it is applicable to many different tasks and robust to various batch sizes without any sensitive hyper-parameter.

(2) SSN is trained using a novel SparsestMax function that turns the sparse optimization problem into a simple forward propagation of a deep network. SparsestMax is an extension of softmax with sparsity guarantee and is designed to be a general technique to learn one-hot distribution. We provide its geometry interpretations compared to its counterparts such as softmax and sparsemax~\cite{MartinsA16}.
%
%
%

(3) SSN is demonstrated in multiple computer vision tasks including image classification in ImageNet \cite{russakovsky2015imagenet}, semantic segmentation in Cityscapes \cite{cordts2016cityscapes} and ADE20K \cite{zhou2017scene}, and action recognition in Kinetics \cite{kay2017kinetics}.
Systematic experiments show that SSN with SparsestMax achieves comparable or better performance than the other normalization methods.
%


\section{Sparse Switchable Normalization (SSN)}
\label{sec:second}

This section introduces SSN and SparsestMax.

\subsection{Formulation of SSN}
\label{sec:SSN}

We formulate SSN as
%
%
\begin{eqnarray}\label{eqn:sn}
&&\hspace{25pt}\hat{h}_{ncij}=\gamma\frac{h_{ncij}-\sum_{k=1}^{|\Omega|}p_k\mu_k}{\sqrt{\sum_{k=1}^{|\Omega|}p_k^\prime\sigma_k^2+\epsilon}}+\beta, \\\nonumber
&&\mathrm{ s.t.}~~\sum_{k=1}^{|\Omega|}p_k=1,~~\sum_{k=1}^{|\Omega|}p_k^\prime=1,~~\forall p_k, p_k^\prime \in\{0,1\}
\end{eqnarray}
where $h_{ncij}$ and $\hat{h}_{ncij}$ indicate a hidden pixel before and after normalization. The subscripts represent a pixel $(i,j)$ in the $c$-th channel of the $n$-th sample in a minibatch.
$\gamma$ and $\beta$ are a scale and a shift parameter respectively.
$\Omega=\{\mathrm{IN},\mathrm{BN},\mathrm{LN}\}$ is a set of normalizers.
%
$\mu_{k}$ and $\sigma_{k}^2$ are their means and variances, where $k\in\{1,2,3\}$ corresponds to different normalizers.
Moreover, $p_k$ and $p^\prime_k$ are importance ratios of mean and variance respectively.
We denote $\p=(p_1,p_2,p_3)$ and $\p^\prime=(p_1^\prime,p_2^\prime,p_3^\prime)$ as two vectors of ratios.

According to Eqn.\eqref{eqn:sn}, SSN is a normalizer with three constraints including $\|\p\|_1 = 1 $, $\|\p^\prime\|_1 = 1$, and for all $p_k,p_k^\prime\in\{0,1\}$.
These constraints encourage SSN to choose a single normalizer from $\Omega$ for each normalization layer.
If the sparse constraint $\forall p_k,p_k^\prime\in\{0,1\}$ is relaxed to a soft constraint $\forall p_k,p_k^\prime\in(0,1)$, SSN degrades SN \cite{luo2018differentiable}.
For example, the importance ratios $\p$ in SN can be learned using $\p=\soft(\z)$, where $\z$ are the learnable control parameters of a softmax function\footnote{The softmax function is defined by $p_k=\soft_k(\z)=\exp (z_k)/\sum_{k=1}^{|\Omega|}\exp (z_k)$.} and $\z$ can be optimized using back-propagation (BP). 
Such slackness has been extensively employed in existing works 
\cite{jang2016categorical,liu2018darts}.
%
However, softmax does not satisfy the sparse constraint in SSN.

\begin{figure}
\begin{center}
\includegraphics[width=1.0\linewidth]{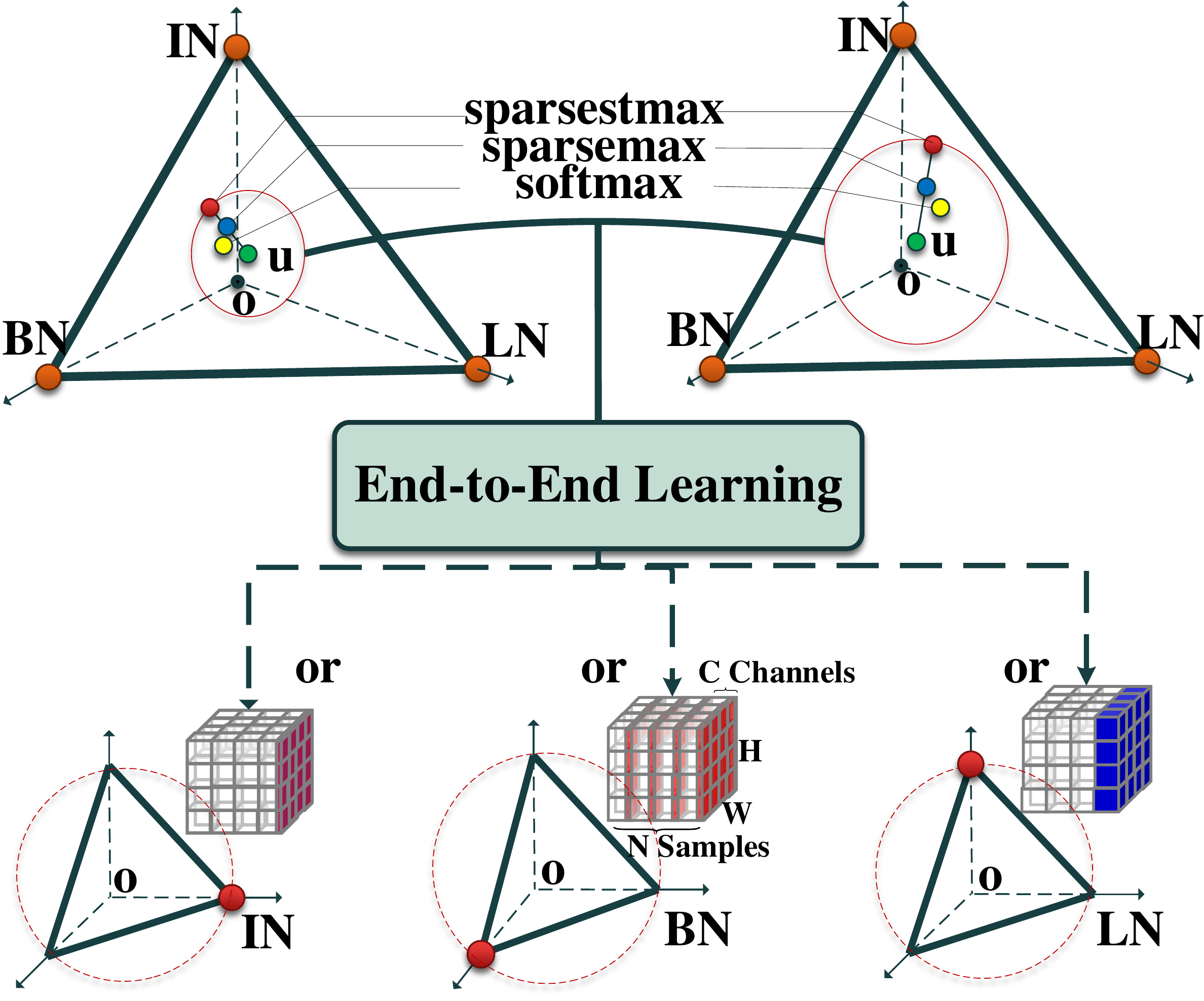}
\end{center}
   \caption{\textbf{Comparisons of softmax, sparsemax and SparsestMax.} 
    $\mathbf{O}$ is the origin of $\R^3$. The regular triangle denotes a 2-D simplex $\triangle^2$ embedded into $\R^3$. $\uu$ is the center of the simplex. The cubes represent feature maps whose dimension is $N\times C\times H\times W$.
   We represent IN, BN and LN by coloring different dimensions of those cubes. Each vertex represents one of three normalizers. As shown in the upper plot, output of softmax is closer to $\uu$ than  sparsemax and SparsestMax. SparsestMax makes important ratios converge to one of vertices of the simplex in an end-to-end manner,  selecting only one normalizer from these three normalization methods.}
\label{fig:figure1}
\end{figure}

%
\textbf{Requirements.} Let $\p=f(\z)$ be a function to learn $\p$ in SSN. Before presenting its formulation, we introduce four requirements of $f(\z)$ in order to make SSN effective and easy to use as much as possible.
(1) \textbf{\emph{Unit length.}} The $\ell1$ norm of $\p$ is $1$ and for all $p_k\geq0$.
%
(2) \textbf {\emph{Completely sparse ratios.}} $\p$ is completely sparse. In other words, $f(\z)$ is required to return an one-hot vector where only one entry is $1$ and the others are $0$s.
%
(3) \textbf{\emph{Easy to use.}} SSN can be implemented as a module and easily plugged into any network and task. To achieve this, all the constraints of $\p$ have to be satisfied and implemented in a forward pass of a network.
%
This is different from adding $\ell_0$ or $\ell_1$ penalty to a loss function, making model development cumbersome because coefficient of these penalties are often sensitive to batch sizes, network architectures, and tasks.
%
(4) \textbf{\emph{Stability.}} The optimization of $\p$ should be stable, meaning that
%
$f(\cdot)$ should be capable to maintain sparsity in the training phase.
For example, training is difficult if $f(\cdot)$ returns one normalizer in the current step and another one in the next step.
%

\noindent\textbf{Softmax and sparsemax?}
Two related functions are softmax and sparsemax, but they do not satisfy all the above requirements.
Firstly, $\soft(\z)$ is employed in SN \cite{luo2018differentiable}.
%
However, its parameters $\z$ always have full support, that is, $p_k=\soft_k(\z)\neq 0$ where $\soft_k(\cdot)$ indicates the $k$-th element, implying that the selection of normalizers is not sparse in SN.

Secondly, another candidate is sparsemax~\cite{MartinsA16} that extends softmax to produces a sparse distribution.
The $\sparse(\z)$ projects $\z$ to its closest point $\p$ on a ($K$-1)-dimensional simplex by minimizing the Euclidean distance between $\p$ and $\z$,
\begin{equation}\label{eqn:sparsemax}
\sparse(\z):=\underset{\p\in\triangle^{K-1}}{\mathrm{argmin}}\left\|\p-\z\right\|_2^2,
\end{equation}
where $\triangle^{K-1}$ denotes a ($K$-1)-D simplex that is a convex polyhedron containing $K$ vertices.
%
%
We have $\triangle^{K-1}:=\{\p \in \R^K|\1\tran \p=1,\p\geq \mathbf{0}\}$ where $\1$ is a vector of ones.
%
For example, when $K=3$, $\triangle^{2}$ represents a 2-D simplex that is 
a regular triangle.
The vertices of the triangle indicate BN, IN, and LN respectively as shown in Fig.\ref{fig:figure1}.
%

By comparing softmax and sparsemax on the top of Fig.\ref{fig:figure1} when $\z$ is the same, the output $\p$ of softmax yellow dot is closer to $\uu$ (center of the simplex) than that of sparsemax blue dot. In other words, sparsemax produces $\p$ that are closer to the boundary of the simplex than softmax, implying that sparsemax produces more sparse ratios than softmax. Take  $\z=(0.8,0.6,0.1)$ as an example,  $\soft(\z)=(0.43,0.35,0.22)$ while $\sparse(\z)=(0.6,0.4,0)$,  showing that sparsemax is likely to make some elements of $\p$ be zero. For the compactness of this work, we provide evaluation of the sparsemax \cite{held1974validation,MartinsA16} in section A of Appendix.
%
%
However, completely sparse ratios cannot be guaranteed because every point on the simplex could be a solution of Eqn.\eqref{eqn:sparsemax}.
%


\subsection{SparsestMax}

To satisfy all the constraints as discussed above, we introduce {\textit{SparsestMax}}, which is a novel sparse version of the softmax function.
%
The SparsestMax function is defined by
\begin{equation}\label{eqn:SparsestMax}
\spest(\z;r):=\underset{\p\in\triangle^{K-1}_r}{\mathrm{argmin}}\left\|\p-\z\right\|_2^2,
\end{equation}
where $\triangle^{K-1}_r:=\{\p \in \R^K|\1\tran \p=1, \left\|\p-\uu\right\|_2\geq r, \p\geq \mathbf{0}\}$ is a simplex with a circular constraint $\left\|\p-\uu\right\|_2\geq r, \1\tran \p=1$. Here $\uu=\frac{1}{K}\1$ is the center of the simplex and $\1$ is a vector of ones, and $r$ is radius of the circle.

Compared to sparsemax, SparsestMax introduces a circular constraint $\left\|\p-\uu\right\|_2\geq r ,\, \1\tran \p=1$ that has an intuitively geometric meaning. Unlike sparsemax where the solution space is $\triangle^{K-1}$, the solution space of SparsestMax is a circle with center $\uu$ and radius $r$  excluded from a simplex. 
%
%

In order to satisfy the completely sparse requirement, we linearly increase $r$ from zero to $r_c$ in the training phase. $r_c$ is the radius of a circumcircle of the simplex.
To understand the important role of $r$, we emphasize two cases.
%
%
When $r\leq \left\|\p_0-\uu\right\|_2$, where $\p_0$ is the output of sparsemax, then $\p_0$ is also the solution of Eqn.(\ref{eqn:SparsestMax}) because $\p_0$ satisfies the circular constraint. 
When $r=r_c$, 
the solution space of Eqn.(\ref{eqn:SparsestMax}) contains only $K$ vertices of the simplex, making $\spest(\z;r_c)$ completely sparse.
%

\begin{figure}[t]
\begin{center}
\includegraphics[width=1.0\linewidth]{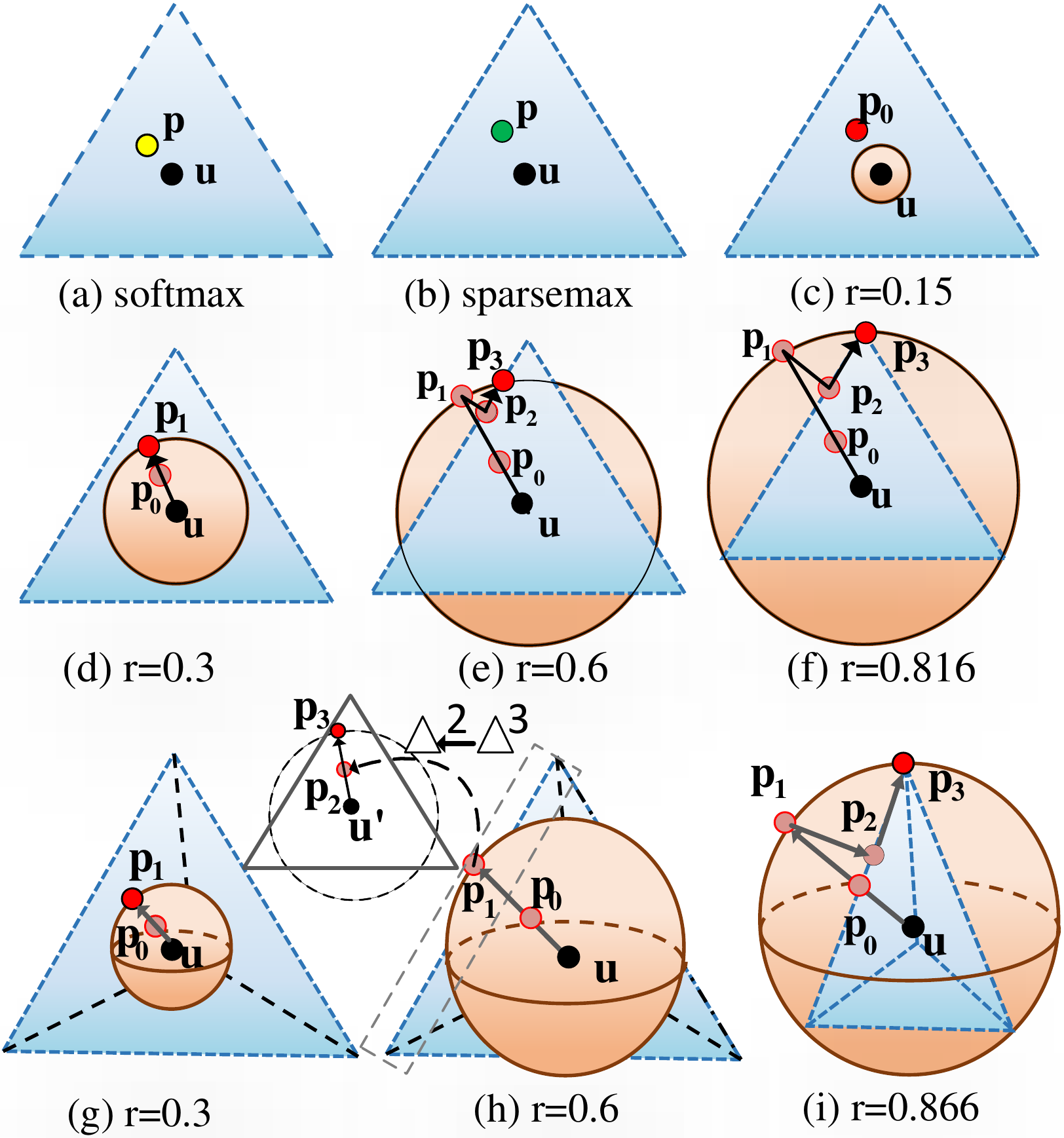}
\end{center}
   \caption{Illustration of (a) softmax, (b) sparsemax, (c-f) SparsestMax when $K=3$ and (g-i) SparsestMax when $K=4$. $\uu$ is the center of the simplex. $\uu=(\frac{1}{3},\frac{1}{3},\frac{1}{3})$ for $K=3$ and $\uu=(\frac{1}{4},\frac{1}{4},\frac{1}{4},\frac{1}{4})$ for $K=4$. Given $\z=(0.5,0.3,0.2)$, (a) and (b) show that the outputs of softmax and sparsemax are  $\p=(0.39,0.32,0.29)$ and $\p=(0.5,0.3,0.2)$ respectively. (c-f) show that the results of SparsestMax for $r=0.15, 0.3, 0.6$ and $0.816$ are $\p_0=(0.5,0.3,0.2)$, $\p_1=(0.56,0.29,0.15)$, $\p_3=(0.81,0.19,0)$ and $\p_3=(1,0,0)$ respectively when $K=3$, a concrete calculation is given in \textbf{Stage 1-4}. When $K=4$, given $\z=(0.3,0.25,0.23,0.22)$, the outputs of (g-i) are $\p_1=(0.49,0.25,0.15,0.11)$, $\p_3=(0.75,0.23,0.02,0)$ and $\p_3=(1,0,0,0)$ for $r=0.3, 0.6$ and $0.866$ respectively. All $\p_2$s are acquired by $\p_2=\sparse (\p_1)$. (e) and (f) show that when $\p_1$ is outside of the simplex $\triangle^{K-1}$, then projection space reduces to $\triangle^{K-2}$ for $K=3$ and $K=4$.}
\label{fig:figure2}
\end{figure}

\textbf{An example.} Fig.\ref{fig:figure2}(a-f) illustrate a concrete example in the case of $K=3$ and $\z=(0.5,0.3,0.2)$.
We can see that the output of softmax is more uniform than sparsemax and SparsestMax, and SparsestMax produces increasingly sparse output as $r$ grows.
With radius $r$ gradually increasing in the training phase, the computations of SparsestMax are discussed as below.
%


\textbf{Stage 1.} As shown in Fig.\ref{fig:figure2}(b,c), the solution of sparsemax is  $\p_0=(0.5,0.3,0.2)$ given $\z=(0.5,0.3,0.2)$. When $r=0.15$, 
$\p_0$ satisfies the constraint $\left\|\p_0-\uu\right\|_2 \geq r$. Therefore, $\p_0$ is also the solution of SparsestMax.
%
In this case, SparsestMax is computed the same as sparsemax to return the optimal ratios.
%

\textbf{Stage 2.} As illustrated in Fig.\ref{fig:figure2}(d), when $r$ increases to $0.3$ and thus $\left\|\p_0-\uu\right\|_2 < r$ when $\p_0=(0.5,0.3,0.2)$, it implies that the circular constraint is not satisfied.
%
In this case, SparsestMax returns the point $\p_1$ on the circle, which is computed by projecting $\p_0$ to the face of circle, that is, $\p_1=r\frac{\p_0-\uu}{\left\|\p_0-\uu\right\|_2}+\uu=(0.56,0.39,0.15)$ as the output.
%
%

\textbf{Stage 3.} As shown in Fig.\ref{fig:figure2}(e), when $r=0.6$, $\p_1$ moves out of the simplex. In this case, $\p_1$ is projected back to the closest point on the simplex, that is $\p_2$, which is then pushed to $\p_3$ by the SparsestMax function using
%
\begin{equation}\label{eqn:reprjball}
\p_3=r^\prime\frac{\p_2-\uup}{\left\|\p_2-\uup\right\|_2}+\uup,
\end{equation}
where $u_i^\prime =\max\{\frac{(\p_1)_i}{2},0\},\, i=1,2,3$, $\p_2=\sparse (\p_1)$ and $r^\prime=\sqrt{r^2-\left\|\uu-\uup\right\|_2^2}$. 
In fact, $\p_2$ lies on $\triangle^1$, $\uu^\prime$ is the center of $\triangle^1$ and $\triangle^1$ is one of the three edges of $\triangle^2$. Eqn.(\ref{eqn:reprjball}) represents the projection from $\p_2$ to $\p_3$.
We have $\p_3=(0.81,0.19,0)$ as the output. It is noteworthy that when $\p_1$ is out of the simplex, $\p_3$ is a point of intersection of the simplex and the circle, and $\p_3$ can be determined by sorting $\p_0$. In this way, Eqn.(\ref{eqn:reprjball}) can be equivalently replaced by argmax function. However,  Eqn.(\ref{eqn:reprjball}) shows great advantage on differentiable learning of parameter $\z$ when $K>3$.
%
%
%

\textbf{Stage 4.} As shown in Fig.\ref{fig:figure2}(f), the circle becomes the circumcircle of the simplex when $r=r_c=0.816$ for K =3, $\p_3$ moves to one of the three vertices. This vertex would be the closest point to $\p_0$. In this case, we have $\p_3=(1,0,0)$ as the output.

%

\textbf{Implementation.} 
In fact, Eqn.(\ref{eqn:SparsestMax}) is an optimization problem with both linear and nonlinear constraints.
The above four stages can be rigorously derived from KKT conditions of the optimization problem.
The concrete evaluation procedure of $\spest$ in case where $K=3$ is presented in Algorithm \ref{alg:algone}.
We see that runtime of Algorithm \ref{alg:algone} mainly depends on the evaluation of sparsemax \cite{van2008probing} (line 1). 
As for SSN, we adopt a $\mathcal{O}(K\mathrm{log}\,K)$ algorithm \cite{held1974validation} to evaluate sparsemax. 
SparsestMax can be easily implemented using popular deep learning frameworks such as PyTorch~\cite{paszke2017automatic}.


%

\subsection{Discussions}

\textbf{Properties of SparsestMax.}
The SparsestMax function satisfies all four requirements discussed before.
Since the radius $r$ increases from 0 to $r_c$ as training progresses,
the solution space of Eqn.\ref{eqn:SparsestMax} shrinks to three vertices of the simplex, returning ratios as a one-hot vector.
The first two requirements are guaranteed until training converged.

%
%

For the third requirement, the SparsestMax is performed in a single forward pass of a deep network, instead of introducing an additional sparse regularization term to the loss function, where strength of the regularization is difficult to tune.

\textbf{Stability of Sparsity.} We explain that training SSN with SparsestMax is stable, satisfying the fourth requirement.
In general, once $p_k=\spest_k(\z;r)=0$ for each $k$, derivative of the loss function wrt. $z_k$ is zero using chain rules as provided in section B of Appendix.
This property explicitly reveals that 
%
once an element of $\p$ becomes 0, it will not `wake up' in the succeeding training phase, which has great advantage of maintaining sparsity in training.

We examine the above property for different stages as discussed before.
Here, we denote $\mathbf{p}-\mathbf{u}$ and $\left\|\mathbf{p}-\mathbf{u}\right\|_2$ as  `\textit{sparse direction}' and `\textit{sparse distance}' respectively.
The situation when $p_k = 0$ only occurs on stage 1 and stage 3.
In stage 1, SparsestMax becomes sparsemax \cite{MartinsA16}, which indicates that if $p_k=0$, the $k$-th component in $\p$ is much less important than the others. Therefore, stopping learning $p_k$ is reasonable.
In stage 3, $p_k=0$ occurs when $\p_0$ moves to $\p_1$ and then $\p_2$. 
In this case, we claim that $\p_1$ has learned a good sparse direction before it moves out of the simplex.
To see this, when $\left\|\p_0-\uu\right\|_2<r,\, \p_1\geq \mathbf{0}$, let $\g_1$ be the gradients of the loss function with respect to $\p_1$ during back-propagation.
We can compute $g_0^d$ that is the directional derivative of the loss at $\p_0$ in the direction $\p_0-\uu$. We have
\begin{equation}\label{eqn:dirg}
	\begin{split}
	g_0^d &=\left(\frac{\partial \p_1}{\partial \p_0}\right)		\tran \g_1(\p_0-\uu)\\
	&=\g_1\tran\frac{\left\|\p_0-\uu\right\|^2-(\p_0-\uu)			(\p_0-\uu)\tran}{\left\|\p_0-\uu\right\|^{\frac{5}{2}}}			(\p_0-\uu)\\
	&=0.
	\end{split}
\end{equation}

Eqn.(\ref{eqn:dirg}) suggests that SGD would learn the sparse direction regardless of the sparse distance.
In other words, the importance ratios in SSN do not need to learn the sparse distance. They focus on updating the sparse direction to regulate the relative magnitudes of IN, BN, and LN in each training step.
This property intuitively reduces the difficulty when training the important ratios. 



\begin{algorithm}[t]
	\caption{SSN with SparsestMax when $K=3$.}
	\label{alg:algone}
	\begin{algorithmic}[1]
		\Require
		 $\z, \z^\prime, \uu, r, \mu_k, \sigma_k^2$
		\Comment{$r$ increases from zero to $r_c$ in the training stage; $\mu_k$ and $\sigma_k$ denote means and variances from different normalizers, $k\in\{1,2,3\}$}
		\Ensure
		$\mu, \sigma^2$
		\Comment{mean and variance in SSN}
		\State  $\p_0=\sparse (\z)$
		\State  if $\left\|\p_0-\uu\right\|_2 \geq r$ \, then
		\State  \quad $\p=\p_0$
		\State  else $\p_1=r\frac{\p_0-\uu}{\left\|\p_0-\uu\right\|_2}+\uu$
		\State  \quad if $\p_1\geq \mathbf{0}$, \, then
		\State	\qquad $\p=\p_1$
		\State  \quad else compute $\uup, r^\prime$ and $\p_2$
		\Comment{see Stage 3}
		\State  \qquad $\p=r^\prime\frac{\p_2-\uup}{\left\|\p_2-\uup\right\|_2}+\uup$
		\State  \quad  end\, if
		\State  end\,if\\
		\Return $\mu=\sum_{k=1}^{3}p_k\mu_k, \sigma^2=\sum_{k=1}^{3}p_k^\prime\sigma_k^2$
		\Comment{$\p^\prime$ is computed the same as $\p$}
	\end{algorithmic}
\end{algorithm}

%



\textbf{Efficient Computations.} Let $L$ be the total number of normalization layers of a deep network. In training phase, computational complexity of SparsestMax in the entire network is $\mathcal{O}(LK\mathrm{log} K)$, which is comparable to softmax $\mathcal{O}(LK)$ in SN when $K=3$.
However, SSN learns a completely sparse selection of normalizers, making it faster than SN in testing phase.
Unlike SN that needs to estimate statistics of IN, BN, and LN in every normalization layers, SSN computes statistics for only one normalizer.
On the other hand, we can turn BN in SSN into a linear transformation and then merge it into the previous convolution layer, which reduces computations.

\textbf{Extend to $\mathbf{K=n}$.}
%
SparsestMax can be generalized to case where $K=n$. As discussed before,  it results in a one-hot vector under the guidance of a increasing circle. SparsestMax works by inheriting good merit of sparsemax and learning a good sparse direction.
By repeating this step, the projection space degenerates, ultimately leading to a one-hot distribution. Fig.\ref{fig:figure2} (g-i) visualizes Stage 2-4 when $K=4$. We list skeleton pseudo-code for SparsestMax when $K=n$ in Algorithm \ref{algsec}

\section{Relation with Previous Work}

As one of the most significant components in deep neural networks, normalization technique~\cite{ioffe2015batch,ba2016layer,ulyanov2017instance,wu2018group,luo2018differentiable} has achieved much attention in recent years.
These methods can be categorized into two groups:
methods normalizing activation over feature space such as \cite{ioffe2015batch,ba2016layer,ulyanov2017instance,wu2018group}
and methods normalizing weights over the parameter space like \cite{salimans2016weight, miyato2018spectral}.
All of them show that normalization methods make great contribution to stabilizing the training and boosting the performance of DNN.
Recent study of IBN~\cite{pan2018two} shows that the hybrid of multiple normalizers in the neural networks can greatly strengthen the generalization ability of DNN.
A more general case named Switchable Normalization (SN)~\cite{luo2018differentiable} is also proposed to select different normalizer combinations for different normalization layers.
Inspired by this work, we propose \textit{\textbf{SSN}} where the importance ratios are constrained to be completely sparse, while inheriting  all benefits from SN at the same time.
Moreover, the one-hot output of importance ratios alleviates overfitting in the training stage and removes the redundant computations in the inference stage.

Other work focusing on the sparsity of parameters in DNN is also related to this article.
In \cite{scardapane2017group}, group Lasso penalty is adopted to impose group-level sparsity on network’s connections. But this work can hardly satisfy our standardization constraints, \ie sum of the importance ratios in each layer equals one.
Bayesian compression~\cite{louizos2017learning} includes a set of non-negative stochastic gates to determine which weight is zero, making re-parameterized $\ell_0$ penalty differentiable. However, such regularization term makes the model less accurate if applied to our setting where required $\ell_0$ norm is exactly equal to one.
Alternatively, sparsemax that preserves most of the attractive properties of softmax is proposed in~\cite{MartinsA16} to generate sparse distribution, but this distribution is usually not completely sparse.
This paper introduces \textit{SparsestMax}, which adds a circular constraint on sparsemax to achieve the goal of SSN.
It learns the sparse direction regardless of sparse distance in the training phase,
and guarantees to activate only one control parameter.
It can be embedded as a general component to any end-to-end training architectures to learn one-hot distribution.
%
\begin{algorithm}[t]
	\caption{SparsestMax for $K=n$}
	\label{algsec}
	\begin{algorithmic}[1]
	\Require
	 $\z, \, \uu, \, r$
	\Ensure
	$\p=\spest (\z,r,\uu)$
	\State $\p_0=\sparse (\z)$
	\State if $\left\|\p_0-\uu\right\|_2 \geq r$ \, then
	\State \quad $\p=\p_0$
	\State else $\p_1=r\frac{\p_0-\uu}{\left\|\p_0-\uu\right\|_2}+\uu$
	\State \quad if $\p_1\geq \0$, \, then
	
	\State \qquad $\p=\p_1$
	\State \quad else compute $\uup, r^\prime$ and $\p_2$
	\Comment{see Stage 3}
	\State \qquad $\z=\p_2$, $\p=\spest (\z,r^\prime,\uup)$
	\State \quad  end\, if
	\State end\,if\\
	\Return $\p$
	\end{algorithmic}
\end{algorithm}
\section{Experiments}\label{sec:experiment}
In this section, we apply SSN to several benchmarks including image classification, semantic segmentation and action recognition. We show its advantages in both performance and inference speed comparing to existing normalization methods.

\subsection{Image Classification in ImageNet}
In our experiments, we first evaluate SSN in the ImageNet classification dataset~\cite{russakovsky2015imagenet}, which has 1.28M training images and 50k validation images with 1000 categories. All classification results are evaluated on the 224$\times$224 pixels center crop of images in validation set, whose short sides are rescaled to 256 pixels.

\textbf{Implementation details.} All models are trained using 8 GPUs and here we denote batch sizes as the number of images on one single GPU and the mean and variance of BN are calculated within each GPU. For convolution layers, we follow the initialization method used by \cite{he2016deep}. Following~\cite{goyal2017accurate}, we initialize $\gamma$ to 1 for last normalization layers in each residual block and use 0 to initialize all other $\gamma$. Learnable control parameters $\z$ in SSN are initialized as 1. SGD with momentum is used for all parameters, while the learning rate of $\z$ is 1/10 of other parameters. We also apply a weight decay of 0.0001 to all parameters except $\z$. We train all the models for 100 epochs and decrease the learning rate by 10$\times$ at 30, 60 and 90 epochs. By default, our hyperparameter radius $r$ used as circular constraint increases from 0 to 1 linearly during the whole training process, and $\z$ will stop updating once its related importance ratio becomes completely sparse.

\begin{figure}
\begin{center}
\includegraphics[width=0.9\linewidth]{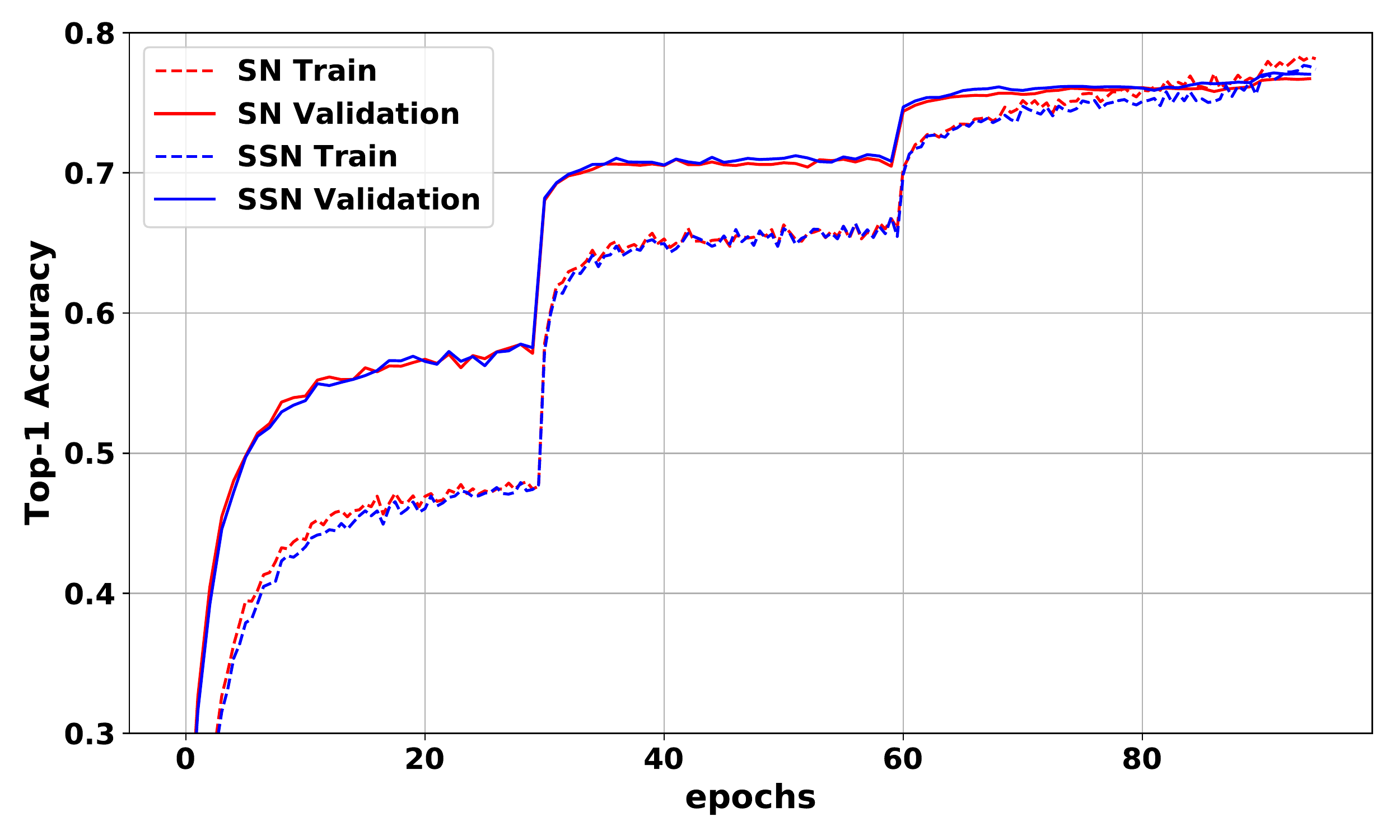}
\end{center}
\caption{\textbf{Training and validation curves of SN and SSN} with a batch size of 32 images/GPU.}
\label{fig:SSNcurve}
\end{figure}

\textbf{Comparison with other normalization methods.} We evaluate all normalization methods using ResNet-50~\cite{he2016deep} with a regular batch size of 32 images per GPU. Table~\ref{tab:different_norms} shows that IN and LN achieve 71.6\% and 74.7\% top-1 accuracy respectively, indicating they are unsuitable for image classification task. BN works quite well in this setting, getting 76.4\% top-1 accuracy. SN combines the advantages of IN, LN and BN and outperforms BN by 0.5\%. Different from SN, SSN selects exactly one normalizer for each normalization layer, introducing stronger regularization and outperforming SN by 0.3\%. Fig.\ref{fig:SSNcurve} shows that SSN has lower training accuracy than SN while maintains even higher validation accuracy.

\begin{table}
\begin{center}
\begin{tabular}{c | c c c c c c c c}
\hline
& IN & LN & BN & GN & SN & SSN \\
\hline
top-1 & 71.6 & 74.7 & 76.4 & 75.9 & 76.9 & \textbf{77.2} \\
$\Delta$ \vs BN & -4.8 & -1.7 & - & -0.5 & 0.5 & \textbf{0.8} \\
\hline
\end{tabular}
\end{center}
\caption{\textbf{Comparisons of top-1 accuracy(\%)} of ResNet-50 in ImageNet validation set. All models are trained with a batch size of 32 images/GPU. The second row shows the accuracy differences between BN and other normalization methods.}
\label{tab:different_norms}
\end{table}


%

%

\begin{figure}
\begin{center}
\includegraphics[width=1\linewidth]{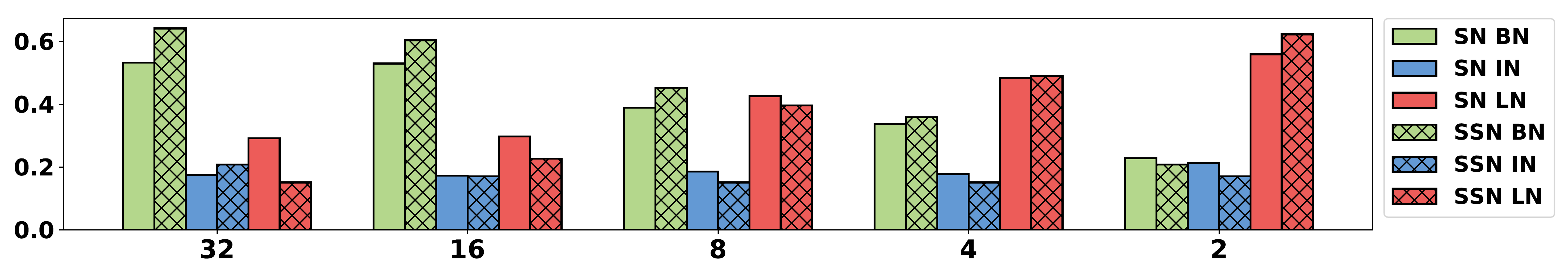}
(a) importance ratios distribution of $\mu$
\includegraphics[width=1\linewidth]{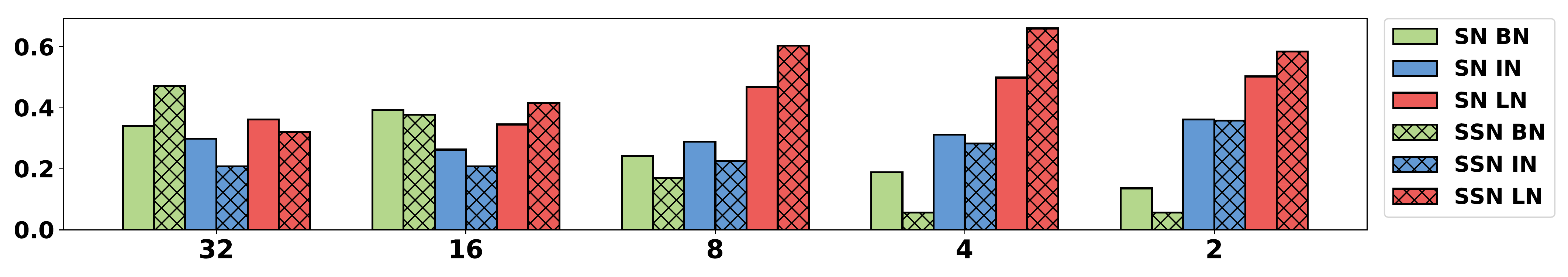}
(b) importance ratios distribution of $\sigma$
\end{center}
\caption{\textbf{Comparisons of importance ratios distribution} between SN and SSN. The model here is ResNet-50 with different batch sizes. (a) visualizes the importance ratios distribution of mean and (b) shows the result of variance. The x-axis denotes the batch size. SSN distribution are shaded.}
\label{fig:SSN-normselect}
\end{figure}

\textbf{Different batch sizes.} For the training of different batch sizes, we adopt the learning rate scaling rule from \cite{goyal2017accurate}, as the initial learning rate is 0.1 for the batch size of 32, and 0.1N/32 for a batch size of N. The performance of BN decreases from $76.4\%$ to $65.3\%$ when the batch size decreases from 32 to 2 because of the larger uncertainty of statistics. While GN and SN are less sensitive to batch size, SSN achieves better performance than these two methods and outperforms them in all batch size settings, indicating SSN is robust to batch size. The top-1 accuracies are reported in Table~\ref{tab:different_batchsize}. In Fig.\ref{fig:SSN-normselect}, we visualize the normalizer selection distribution of SSN in different batch sizes. Our results show that the network would prefer BN in larger batch size while LN in smaller batch size. We can also observe that the importance ratio distribution is generally different between $\mu$ and $\sigma$ which is consistent with study in \cite{luo2018understanding,teye2018bayesian}. At the same time, SSN would have a more distinct importance ratio distribution than SN.

\begin{table}
\begin{center}
\begin{tabular}{c|c c c c c}
\hline
batch size & 32 & 16 & 8 & 4 & 2 \\
\hline
BN & 76.4 & 76.3 & 75.2 & 72.7 & 65.3 \\
GN & 75.9 & 75.8 & 76.0 & 75.8 & \textbf{75.9} \\
SN & 76.9 & 76.7 & 76.7 & 75.9 & 75.6 \\
SSN & \textbf{77.2} & \textbf{77.0} & \textbf{76.8} & \textbf{76.1} & \textbf{75.9} \\
\hline
\end{tabular}
\end{center}
\caption{\textbf{Top-1 accuracy in different batch sizes.} We show ResNet-50's validation accuracy in ImageNet. SSN achieves higher performance in all batch size settings.}
\label{tab:different_batchsize}
\end{table}

\textbf{Fast inference.} Different from SN, SSN only need to select one normalizer in each normalization layer, saving lots of computations and graphic memories.  We test our inference speed using the batch size of 32 images on a single GTX 1080. For fair comparison, we implement all normalization layers in PyTorch.
All BN operations are merged into previous convolution operations. As showed in Table~\ref{tab:throughput}, 
BN is the fastest. SSN uses BN, IN, or LN in each layer, being the second fastest. SSN is faster than IN, LN, GN and SN in both ResNet-50 and ResNet-101 backbone. GN is slower than IN because it divides channels into groups. SN soft combines BN, IN, and LN, making it slower than SSN. 

\begin{table}
\begin{center}
\begin{tabular}{ c | c c }
\hline
& ResNet-50 & ResNet-101 \\
\hline
BN & 259.756~$\pm$~2.136 & 157.461~$\pm$~0.482 \\
\hline
IN & 186.238~$\pm$~0.698 & 116.841~$\pm$~0.289 \\
LN & 184.506~$\pm$~0.054 & 115.070~$\pm$~0.028 \\
GN & 183.131~$\pm$~0.277 & 113.332~$\pm$~0.023 \\
SN & 183.509~$\pm$~0.026 & 113.992~$\pm$~0.015 \\
SSN & \textbf{216.254~$\pm$~0.376} & \textbf{133.721~$\pm$~0.106} \\
\hline
\end{tabular}
\end{center}
\caption{\textbf{Throughput (images/second) in inference time} over different normalization methods with ResNet-50 and ResNet-101 as backbone. Larger is better. The mean and standard deviation are calculated over 1000 batches.}
\label{tab:throughput}
\end{table}

\begin{figure*}
\begin{center}
\includegraphics[width=0.9\linewidth]{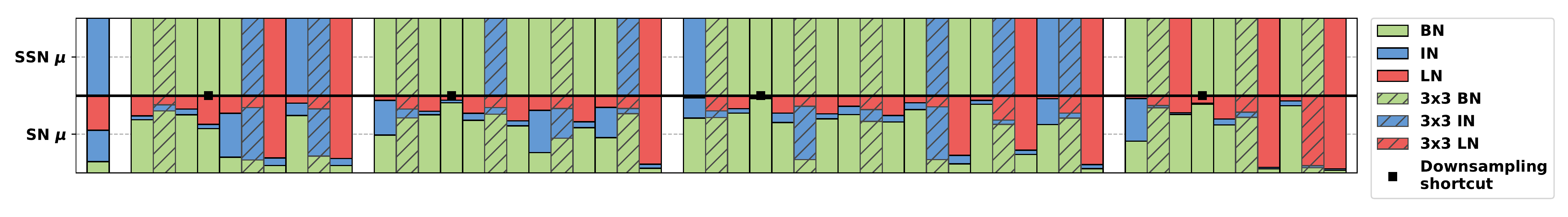}
\includegraphics[width=0.9\linewidth]{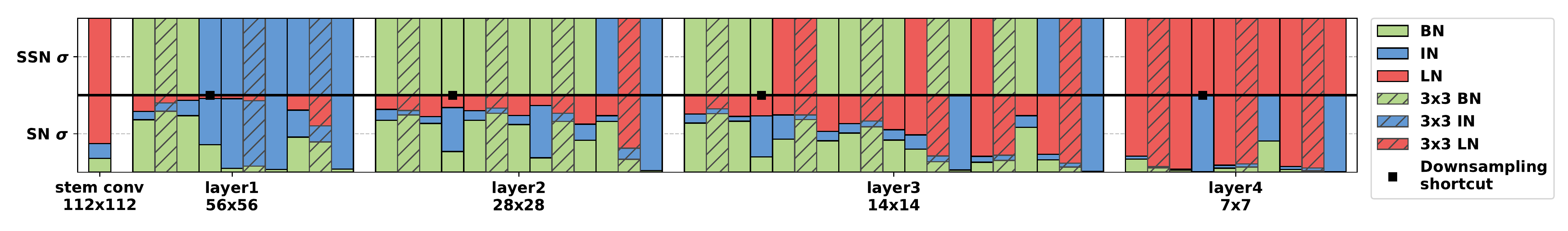}
\end{center}
    \caption{\textbf{Comparison of normalizer selection} between SSN and SN with a batch size of 32 images per GPU. The network we use is ResNet-50, which has 53 normalization layers. The top and bottom plots denote importance ratios of mean and variance respectively. We shade normalizers after 3x3 conv and mark normalizers after downsampling shortcut with ``$\blacksquare$''. }
\label{fig:SSN-SN_layernorm}
\end{figure*}

 \textbf{Comparison of normalizer selection between SSN and SN.} Fig.\ref{fig:SSN-SN_layernorm} compares the breakdown results of normalizer selection between SSN and SN for all normalization layers in ResNet-50 with a batch size of 32. Almost all dominating normalizers in SN are selected by SSN.
 By our analysis, those normalization layers with uniform importance ratios in SN are expected to focus on learning sparse direction in SSN, and converge to a more appropriate normalizer.

\begin{figure}
\begin{center}
\includegraphics[width=0.9\linewidth]{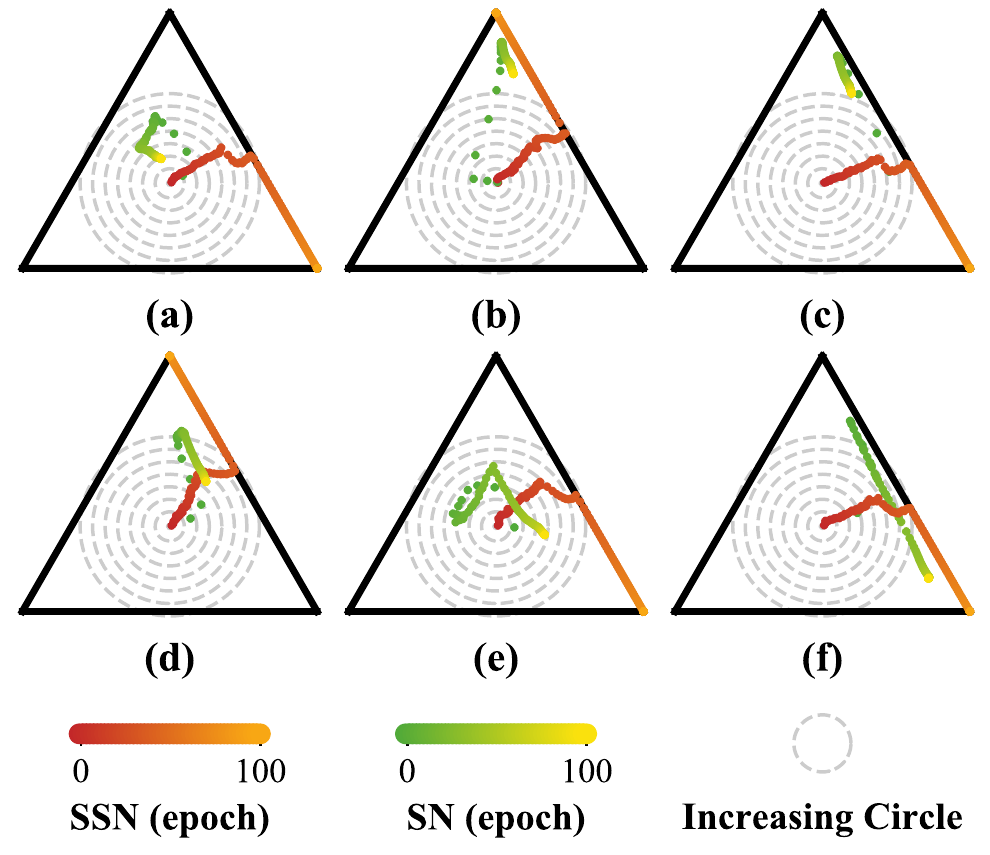}
\end{center}
\caption{\textbf{Comparison of convergence of importance ratios} in some normalization layers across the network. These plots visualize the variance importance ratios in (layer3.0.norm2), (layer3.1.norm1), (layer3.1.norm2), (layer3.2.norm1), (layer3.3.norm1) and (layer3.4.norm1) of ResNet-50 respectively.}
\label{fig:SSN-SN-sparsetraj}
\end{figure}

\textbf{Learning sparse direction.}
 As we mentioned in Eqn.(\ref{eqn:dirg}), SparsestMax focuses on learning sparse direction regardless of sparse distance of importance ratios. To verify this property, we visualize the convergence trajectory of importance ratios of some normalization layers across the network. As shown in Fig.\ref{fig:SSN-SN-sparsetraj}, the importance ratios in SSN make adjustment for their sparse directions under the guidance of an increasing circle in each iteration, and keep the direction stable until completely sparse. While the convergence behavior of those ratios in SN seems to be a bit messy.

 %

\textbf{Insensitiveness to $r$'s increasing schedule.}
SSN has an important hyperparameter $r$, which is the radius of increasing circle. Here we examine that SSN is insensitive to $r$'s increasing schedule. Through our analysis, once the increasing circle is bigger than the inscribed circle of the simplex (i.e. $r>r_i=\sqrt{6}/6$ in the case of three normalizers), the sparse direction is likely to stop updating. In this case, the normalizer selection is determined since the gradients wrt. the control parameters become zero.
Therefore, the time stamp which $r$ reaches $r_i$ matters most in the increasing schedule.
In our default setting, $r$ would increase to $r_i$ at about 41 epoch when training 100 epochs. In our experiment, we make $r$ reach $r_i$ at 40, 50, 60 and 70 epochs respectively. Our result shows that the performance maintains at 77.2$\pm$0.04\%, showing that the schedule contributes little to the final performance.

%





\textbf{One Stage \vs Two Stage.} We use argmax to derive a sparse normalizer architecture from pretrained SN model and compare it with SSN. For comparison, we continue to train argmaxed SN for 20 epochs with an initial learning rate of 0.001 and the cosine annealing learning rate decay schedule. As a result, the sparse structure derived from SN model only reaches 76.8\%, which is not comparable to our one-stage SSN.
In all, SSN obtains sparse structure and shows better performance without introducing additional computation.

\textbf{Four normalizers in $\Omega$.} To evaluate the extensibility of SparsestMax, we introduce GN~\cite{wu2018group} to initial $\Omega$ which contains IN, BN and LN. 
For GN, we use a group number of 32 which is the same as default setting in \cite{wu2018group}. We apply both SN and SSN given the new $\Omega$ to ResNet-50 with a batch size of 32.
In such setting, that SSN obtains higher accuracy 77.3\% than 76.8\% in SN, demonstrating the potential extensibility of SparsestMax in a more generalized scenario.

\subsection{Semantic Segmentation in ADE and Cityscapes}
To investigate generalization ability of SSN in various computer vision tasks, we evaluate SSN in semantic segmentation with two standard benchmarks, \ie ADE20K~\cite{zhou2017scene} and Cityscapes~\cite{cordts2016cityscapes}.
For both of these two datasets, we use 2 samples per GPU.
For fair comparison with SN~\cite{luo2018differentiable}, we also adopted DeepLab~\cite{chen2018deeplab} with ResNet-50 as the backbone network,
where output\_stride=8 and
the last two blocks in the original ResNet contains atrous convolution layers with rate=2 and rate=4, respectively.
Then the bilinear operation is used to upsample the score maps to the size of ground truth.
In the training phase, we use the `poly' learning rate in both two datasets with power=0.9 and the auxiliary loss with the weight 0.4.
The same setting is also used in ~\cite{zhao2017pspnet}.
We compare proposed SSN with Synchronized BN (SyncBN), GN and SN.
For the former three normalization methods, we adopted their pretrained models in ImageNet.
For SSN, we employ SN ImageNet pretrained model~\cite{luo2018differentiable} and use SparsestMax to make the importance ratios completely sparse.
Note that the Synchronized BN is not adopted in both SN and SSN.

For ADE20K, we resize the input image to 450$\times$450 and train for 100,000 iterations with the initial $lr$ 0.02.
For multi-scale testing, we set input\_size=\{300, 400, 500, 600\}.
Table~\ref{tab:segmentation} reports the experiment result in the ADE20K validation set.
SSN outperforms SyncBN and GN with a margin without any bells and whistles in the training phase.
It also achieves 0.2\% higher mIoU than SN in the multi-scale testing.

For Cityscapes, we use random crop with the size 713$\times$713 for all models, and train them with 400 epochs.
The initial $lr$ is 0.01.
The multiple inference scales are \{1.0, 1.25, 1.5, 1.75\}.
According to Table~\ref{tab:segmentation}, SSN performs much better than SyncBN and GN.
It achieves comparable result with SN (75.7~\vs~75.8) in this benchmark.

\begin{table}
\begin{center}
\begin{tabular}{c| c|  c}
\hline
         & ADE20K mIoU$\%$  &  Cityscapes mIoU$\%$  \\
\hline
SyncBN      & 37.7     & 72.7 \\
GN         & 36.3     & 72.2 \\
SN          & 39.1     &\textbf{75.8} \\
SSN         & \textbf{39.3}     & 75.7  \\
\hline
\end{tabular}
\end{center}
\caption{\textbf{Experiment results in ADE20K validation set and cityscapes test set.}
The backbone network is ResNet-50 with dilated convolution layers.
We use mutli-scale inference in the test phase.
SyncBN denotes multi-GPU synchronization of BN.}
\label{tab:segmentation}
\end{table}

\subsection{Action Recognition in Kinetics}
We also apply SSN to action recognition task in Kinetics dataset~\cite{kay2017kinetics}. Here we use Inflated 3D (I3D) convolutional networks~\cite{carreira2017quo} with ResNet-50 as backbone. The network structure and training/validation settings all follow ResNet-50 I3D in \cite{NonLocal2018,wu2018group}. We use 32 frames as input for each video, these frames are sampled sequentially with one-frame gap between each other and randomly resized to [256,320]. Then 224$\times$224 random crop is applied on rescaled frames, and the cropped frames are passed through the network. To evaluate SSN, we use two types of pretrained models here, ResNet-50 SSN with all normalizer selections fixed and ResNet-50 SN with combined normalizers. ResNet-50 SN are trained using SparsestMax to learn sparse normalizer selection in Kinetics. Models are all trained in Kinetics training set using 8 GPUs, and the batch size settings used here are 8 and 4 videos.

During evaluation, for each video we average softmax scores from 10 clips as its final prediction. These clips are sampled evenly from whole video, and each one of them contains 32 frames. The evaluation accuracies in Kinetics validation set are shown in Table~\ref{tab:kinetics}. Both SSN$^1$and SSN$^2$ outperform the results of BN and GN in the batch size of 8 videos per GPU, and SSN$^1$ achieves the highest top-1 accuracy, it's 0.26\% higher than SN and 0.46\% higher than BN. For smaller batch size setting, the performance of SSN lies between SN and GN.
\begin{table}
\begin{center}
\begin{tabular}{c| c c | c c }

\hline
 & \multicolumn{2}{c}{batch=8, length=32} & \multicolumn{2}{|c}{batch=4, length=32} \\

\hline
 & top1 & top5 & top1 & top5 \\
\hline
BN & 73.3 & 90.7 & 72.1 & 90.0 \\
GN & 73.0 & 90.6 & 72.8 & 90.6 \\
SN & 73.5 & \textbf{91.2} & \textbf{73.3} & \textbf{91.2} \\
SSN$^1$ & \textbf{73.8} & \textbf{91.2} & 72.8 & 90.6 \\
SSN$^2$ & 73.4 & 91.1 & 73.0 & \textbf{91.2} \\
\hline
\end{tabular}
\end{center}
\caption{\textbf{Result of ResNet-50 I3D in Kinetics} with different normalization layers and batch sizes. SSN$^1$ is finetuned from ResNet-50 SSN ImageNet pretrained model, and SSN$^2$ is from ResNet-50 SN ImageNet pretrained model.}
\label{tab:kinetics}
\end{table}

\section{Conclusion}
In this work, we propose SSN for both performance boosting and inference acceleration.
SSN inherits all advantages of SN such as robustness to a wide range of batch sizes and applicability to various tasks, while avoiding redundant computations in SN.
This work has demonstrated SSN's superiority in multiple tasks of Computer Vision such as classification and segmentation.
%
To achieve SSN, we propose a novel sparse learning algorithm SparsestMax which turns constrained optimization problem into differentiable feed-forward computation.
We show that SparsestMax can be built as a block for learning one-hot distribution in any deep learning architecture and is expected to be trained end-to-end without any sensitive hyperparameter.
The application of proposed SparsestMax can be a fruitful future research direction.

{\small
\bibliographystyle{ieee}
\bibliography{refb}
}

\clearpage

\renewcommand{\thesection}{\Alph{section}}
\setcounter{section}{0}
{\Large \textbf{Appendices}}

\section{Evaluation for Sparsemax}
The following theorem gives the closed-form solution to the standard sparsemax which is also provided in \cite{MartinsA16, held1974validation}.
\begin{theorem}\label{theorem:sparsemax}
Defining the optimization problem, $\p=\sparse(\z):=\underset{\p\in\triangle^{K-1}}{\mathrm{argmin}}\left\|\p-\z\right\|_2^2$, then solution is of the form:
\begin{equation}\label{eqn:spaesemax}
p_i=\max\{z_i-\tau (\z),0\}
\end{equation}
where $\tau :\R^K \rightarrow \R$ is a function of $\z$. Specifically, given the sort of $\z$,  $z_{(1)}\geq z_{(2)}\geq \cdots \geq z_{(K)}$, define $a(\z)=\max\{k\in[K]|1+kz_{(k)}>\sum_{j\leq k}z_{(j)}\}$ and the support of $\p$ as $S(\z)=\{j\in [K]|p_j>0\}$, then
\begin{equation}\label{eqn:tau}
\tau (\z)=\frac{(\sum_{j\leq a(\z)}z_{(j)})-1}{a(\z)}=\frac{(\sum_{j\in S(\z)}z_{j})-1}{|s(\z)|}
\end{equation}
\end{theorem}
Theorem \ref{theorem:sparsemax} states that all we need for evaluating the sparsemax is to compute threshold $\tau(\z)$
according to the sorted elements of $\z$. Then all elements
above this threshold will be shifted by $\tau(\z)$, and the
others will be set to zero.
For example, when $\z=(0.8,0.6,0.1)$, by Theorem \ref{theorem:sparsemax}, we have $a(\z)=2$ and thus $\tau(\z)=0.2$. Hence, $\sparse(\z)=(0.6,0.4,0)$.

\section{Stability of Sparsity of SparsestMax}
In this section  we focus on case where $K=3$ to explain the stability of sparsity of SparsestMax through its back-propagation. We denote $\p=\spest (\z,r)$ and $\p_0=\sparse (\z)$ given $r\in [0,r_c]$ and consider three cases in Algorithm \ref{alg:algone} when the gradient back-propagates . 

\textbf{Case 1}, if $\left\|\p_0-\uu\right\|_2 \geq r$, we have $\p=\p_0$. Note that $(\p_0)_i=\max\{z_i-\tau(\z),0\}$ where $(\cdot)_i$ denotes $i$-th component of a vector, the gradient of $p_i$ wrt. $z_j$ can be calculated as 
\begin{equation}\label{eq:gradp0}
\frac{\partial p_i}{\partial z_j}=\frac{\partial (p_0)_i}{\partial z_j}=\left\{
\begin{array}{lcl}
\delta_{ij}-\frac{1}{|S(\z)|} & \mathrm{if} \, i,j\in S(\z),\\
0 & \mathrm{else}
\end{array}
\right.
\end{equation}
where $S(\z)=\{i|(\p_0)_i>0,i=1,2,3\}$ and $\delta_{ij}$ is equal to one when $i=j$ and zero otherwise. Eqn.(\ref{eq:gradp0}) shows that if $p_k=0$ then $\frac{\partial p_i}{\partial z_k}=0$ for all $i=1,2,3$ since $k\notin S(\z)$.

\textbf{Case 2}, when $\left\|\p_0-\uu\right\|_2 < r$ and $\p_1>\0$, we have $\p=\p_1$. The condition $\p_1>\0$ implies that $\p_0>\0$. By Eqn.(\ref{eq:gradp0}), the gradient $\frac{\partial (\p_0)_i}{\partial z_j} \neq 0$ for every $i$ and $j$. We obtain
\begin{equation}
\frac{\partial p_i}{\partial z_j} =\sum_{k=1}^3\frac{\partial p_i}{\partial(\p_0)_k}\frac{\partial (\p_0)_k}{\partial z_j}\neq 0,
\end{equation}
which shows that learning of important ratios $\p$ depends on all $z_i,i=1,2,3$. Besides, Equation (\ref{eqn:dirg}) in the paper reveals that SGD would learn the sparse direction regardless of the sparse distance in this case since the directional derivative of the loss at $\p_0$ in the direction $\p_0-\uu$ is equal to zero.

\textbf{Case 3}, when $\left\|\p_0-\uu\right\|_2 < r$ and some elements of $\p_1$ are negative, we have $\p=\p_3$ which is obtained by Eqn.(\ref{eqn:reprjball}). $\p_1$ has negative components, meaning that the corresponding elements of $\p_2$ are exactly equal to zero and so does $\p$.
%
If $(\p_2)_k=0$, then $\frac{\partial (\p_2)_i}{\partial (\p_1)_k} = 0$ for all $i=1,2,3$ since $\p_2=\sparse (\p_1)$.
 
 According to the above discussions, we conclude that the situation when $p_k = 0$ only occurs in Case 1 and 3 which correspond to Stage 1 and 3  in the paper respectively. In addition,  once $p_k = \spest(\z; r) = 0$ for each $k$, the derivative of the loss function wrt. $z_k$ is zero using chain rules.  In  Case 1, SparsestMax has the same output as sparsemax. Hence, $p_k = 0$ implies that the $k$-th component in $\p$ is much less important than those positive components in $\p$. In Case 3, it has been explained in the paper that $\p_1$ has learned a good sparse direction before $p_k$ decreases to zero.

\end{document}